\documentclass[letterpaper, 10 pt, conference]{ieeeconf}  %

\IEEEoverridecommandlockouts                              %

\usepackage{graphicx} %
\usepackage{amsmath} %
\usepackage{amssymb}  %
\usepackage{amsfonts}  %
\usepackage{hyperref}
\usepackage{ifthen}
\usepackage{booktabs}\usepackage{multirow}
\usepackage{verbatim}
\usepackage{graphicx}
\usepackage{xcolor}

\newboolean{useJPEG}
\setboolean{useJPEG}{true}  %
\usepackage{esvect}

\newcommand{\todo}[1]{\textcolor{blue}{\bf #1}}

\newcommand{\etal}{\emph{et al.~}}

\newcommand{\ie}{\emph{i.e.}}
\newcommand{\E}{\mathbb{E}}
\newcommand{\R}{\mathbb{R}}

\title{\LARGE \bf Toward Sim-to-Real Directional Semantic Grasping}
\author{\begin{tabular}{ccccc}Shariq Iqbal$^{1,2}$\thanks{Work was performed while the first author, who is also affiliated with the University of Southern California, was an intern with NVIDIA.  Email: {\tt\footnotesize shariqiqbal2810@gmail.com, \{jtremblay, sbirchfield\}@nvidia.com}}%
& Jonathan Tremblay$^1$ & Thang To$^1$ & Jia Cheng$^1$ & Erik Leitch$^1$\end{tabular}\\ \begin{tabular}{cccc}Andy Campbell$^1$ & Kirby Leung$^1$ & Duncan McKay$^1$ & Stan Birchfield$^1$ \\ \multicolumn{4}{c}{$^1$NVIDIA \hspace{3em} $^2$USC}\end{tabular}}

\begin{document}

\maketitle
\thispagestyle{empty}
\pagestyle{empty}

\begin{abstract}
We address the problem of \emph{directional semantic grasping}, that is, grasping a specific object from a specific direction.
We approach the problem using deep reinforcement learning via a double deep Q-network (DDQN) that learns to map downsampled RGB input images from a wrist-mounted camera to Q-values, which are then translated into Cartesian robot control commands via the cross-entropy method (CEM).
The network is learned entirely on simulated data generated by a custom robot simulator that models both physical reality (contacts) and perceptual quality (high-quality rendering).
The reality gap is bridged using domain randomization.
The system is an example of end-to-end (mapping input monocular RGB images to output Cartesian motor commands) grasping of objects from multiple pre-defined object-centric orientations, such as from the side or top.  
We show promising results in both simulation and the real world, along with some challenges faced and the need for future research in this area.

\end{abstract}

\section{INTRODUCTION}
\label{sec:introduction}

Recent research has made substantial progress in addressing the indiscriminate grasping problem, in which a robot learns to grasp any of several objects from a cluttered bin \cite{mahler2016icra:dex,mahler2017rss:dex,mahler2017corl:binpicking,mahler2018icra:suction}. 	
In such approaches, it does not matter \emph{which} object is grasped, only \emph{that} one or more objects are grasped.  
These methods tend to be restricted to top-down grasping, in which the robot reaches down into a bin using an overhead camera for sensing.

More recent work has addressed the problem of \emph{semantic grasping}, in which a robot learns to grasp a specific type of object---indicated by the user---from a cluttered bin \cite{jang2017corl:endtoend}.  
Although such approaches address the question of \emph{which} object to grasp, they do not address the question of \emph{how} to grasp the objects, {\em i.e.}, the part of the object and direction of the grasp.  
Moreover, these methods are restricted to top-down grasping using overhead sensing.

In this paper, we take the next logical step and address the problem of grasping specific objects from specific directions.  
For many real-world tasks, the manner in which the object is grasped is important.  
To place one object adjacent to another, for example, a top-down grasp may be insufficient, because it would cause the end effector to collide with the other object during placement; rather, a side grasp (on the opposite side) is needed.
In such cases, grasping objects in a specific manner, from a specific direction, is crucial.
We refer to this problem as \emph{directional semantic grasping}.

\begin{figure}
\centering
\begin{tabular}{cc}
Sim & Real \\
\raisebox{1em}{\rotatebox{90}{external view}} 
\includegraphics[width=0.44\columnwidth]{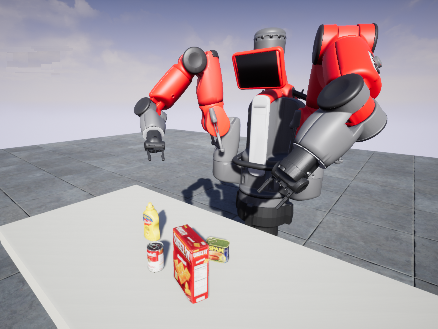} &
\includegraphics[width=0.44\columnwidth]{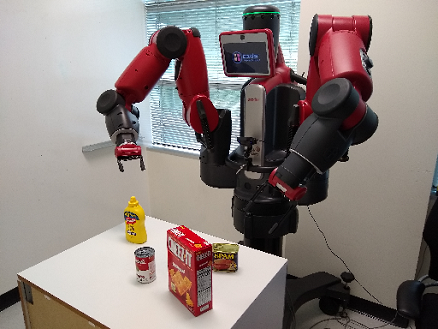} \\
\raisebox{1em}{\rotatebox{90}{gripper camera}} 
\includegraphics[width=0.44\columnwidth]{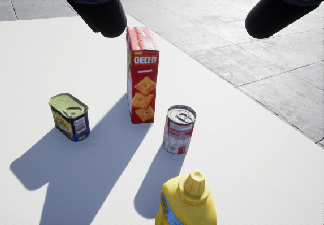} &
\includegraphics[width=0.44\columnwidth]{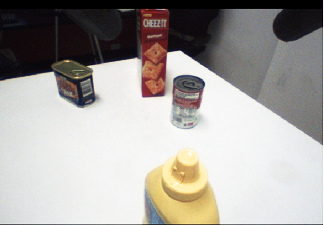}
\end{tabular}
\caption{We have developed a simulator that models physical contact and produces photorealistic imagery.  The goal of this work is to learn a policy in simulation (left) that transfers to the real world (right) for grasping a specific object in a specific way using monocular RGB images from a gripper camera.}
\label{fig:baxter_simtoreal}
\end{figure}

The question we address in this work is whether it is possible to learn an end-to-end policy 
to grasp a specific object from a specific direction using an RGB camera-in-hand.
Further we also investigate whether this policy can be learned using simulation only. 
Simulation has the potential to unlock new robotic learning approaches by generating an almost unlimited amount of training data, essentially for free.
To train a policy that overcomes the \emph{reality gap} problem (\emph{i.e.}, the mismatch between simulated and real distributions),
we have developed a simulation environment with the following design criteria, see Fig.~\ref{fig:baxter_simtoreal}: 
1) it models physical contact, 
2) it generates photorealistic images for training policies based on RGB alone, and 
3) it supports domain randomization to facilitate invariance to various lighting conditions and background textures.

Our contributions are thus:
\begin{itemize}
     \item A deep reinforcement learning (RL) approach to \emph{directional semantic grasping} that learns to grasp specific objects along specific grasp directions in an end-to-end fashion from RGB images from an in-hand camera.
     \item A robotic simulator that generates physically plausible, photorealistic synthetic data, with contact modeling and domain randomization, to train policies that transfer to the real world without a special domain adaptation step.
		\item Promising results on both simulated and real data, along with lessons learned and suggestions for future work.
\end{itemize}

\begin{figure}
\centering
\begin{tabular}{c}
\includegraphics[width=0.9\columnwidth]{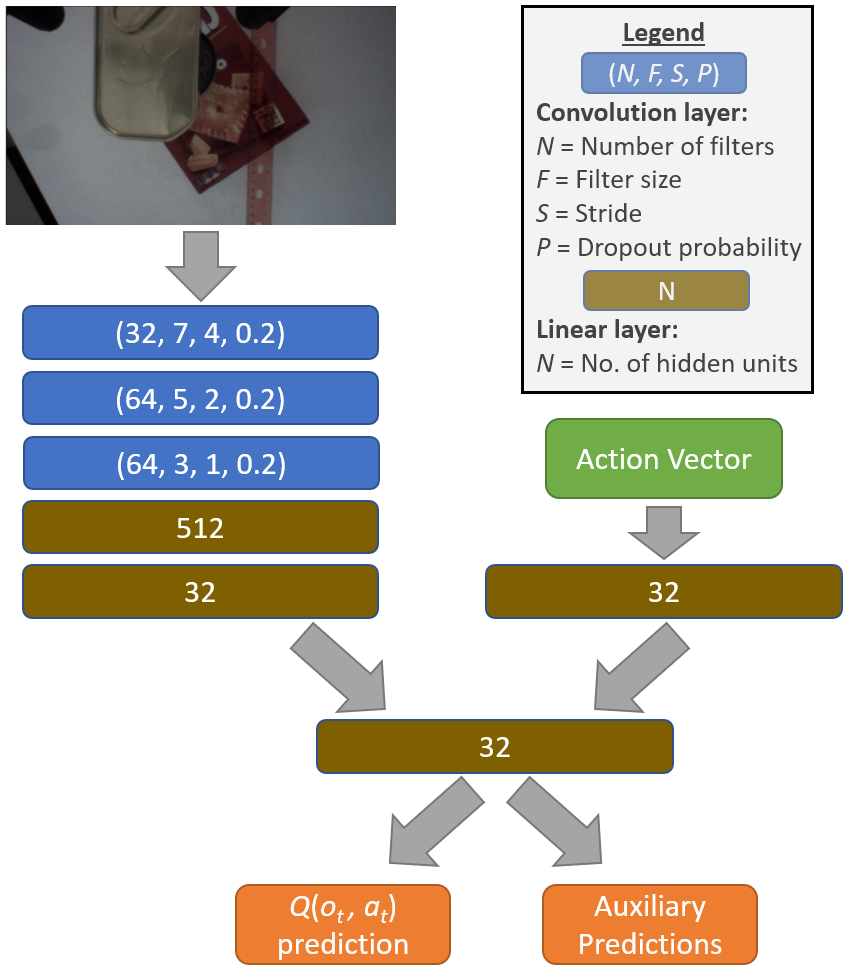}
\end{tabular}
\caption{Detailed network structure:  The RGB image from the gripper camera is embedded by a deep neural network.  This embedding is then concatenated with the action vector (after a single linear layer) to an additional layer to estimate the Q-value, as well as auxiliary targets.  Both the Q-value loss and auxiliary losses are used during training.}
\label{fig:netstruct}
\end{figure}

\section{METHOD}

Adopting the standard reinforcement learning formulation, we model the problem as a partially observable Markov decision process (POMDP) represented as a tuple $(S,O,A,P,r,\gamma)$, where $S$ is the set of states in the environment, $O$ is the set of observations (\ie, downsampled RGB images from the wrist-mounted camera), $A$ is the set of actions (\ie, end-effector-centric 3D Cartesian motion and rotation about the wrist axis), $P:S \times A \times S \rightarrow \R$ is the state transition probability function, $r: S \times A \rightarrow \R$ is the reward function, and $\gamma$ is a discount factor.

The goal of training is to learn a deterministic policy $\pi: O \rightarrow A$ such that taking action $a_t=\pi(o_t)$ at time $t$ maximizes the sum of discounted future rewards from state $s_t$:  $R_t = \sum_{i=t}^\infty \gamma^{i-t}r(s_i,a_i)$.  After taking action $a_t$, the environment transitions from state $s_t$ to state $s_{t+1}$ by sampling from $P$.  The quality of taking action $a_t$, given observation $o_t$ is measured by $Q(o_t,a_t)=\E[R_t|o_t,a_t]$, known as the $Q$-function.

In our environment, each action $a_t=(\delta x_t, \delta y_t, \delta z_t, \delta \phi_t)$ has 4 degrees of freedom, where $(\delta x_t, \delta y_t, \delta z_t) \in [-\Delta_{tran}, \Delta_{tran}]^3$ denotes the translation of the gripper (in the gripper coordinate frame), and $\delta \phi_t \in [-\Delta_{rot},\Delta_{rot}]$ denotes the change in rotation of the gripper about its longitudinal axis; where we set the maximum translation and rotation to $\Delta_{tran}=5$~cm and $\Delta_{rot}=\pi/4$~rad, respectively.  Note that since the action is relative to the gripper, rather than in the world coordinate frame, the system is agnostic to the absolute orientation of the object being grasped.  Each observation $o_t \in \R^{w \times h \times 3}_+$ is an RGB image from the camera in hand, with $w=64, h=40$.

Recent work in reinforcement learning has extended classic approaches to complex observation spaces (e.g., images) \cite{mnih2015nature:dqn} and continuous action spaces (e.g., robot velocities) \cite{lillicrap2015continuous} using deep neural networks as function approximators for $Q(o_t, a_t)$ and/or $\pi(o_t)$. 
In this work, we use Deep Q-learning (DQN)~\cite{mnih2015nature:dqn}, an off-policy, model-free RL algorithm, which only requires learning one network (the Q-function).
Specifically, we use a double deep Q-network (DDQN) \cite{vanhasselt2016aaai:ddqn}, which overcomes an important limitation of DQN (namely, the overestimation of action values) by using separate networks for action selection and action evaluation, copying the weights from one to the other periodically.

As shown in Fig.~\ref{fig:netstruct}, the learned Q-function maps a downsampled $64 \times 40$ RGB image from the camera mounted on the wrist of the robot as well as a continuous 4D value representing an action in the end-effector-centric Cartesian coordinate system to a single scalar value, $Q(o_t, a_t)$.  
Gripper images and actions are processed separately (using a convolutional net with dropout~\cite{srivastava2014dropout} for the former and a single linear layer for the latter) before concatenating their intermediate representations into a single vector that is fed through a linear layer and then split into two output layers: one for predicting Q-values and another for predicting auxiliary targets. Rectified Linear Units (ReLUs) are used for nonlinearity throughout the network. 
The auxiliary targets are heuristically selected features correlated with rewards for improving the learning signal available to the network. We use the following two quantities as auxiliary targets: 1) distance of the gripper to the centroid of the current object, and 2) the rotational offset of the gripper with respect to the object. 
Mean squared error is used as the loss function for both.

Since Q-learning requires selecting the action with the highest Q-value, and since our action space is continuous, this maximum needs to be approximated. 
We adopt the approach of prior work \cite{levine2018ijrr:learning,jang2017corl:endtoend,quillen2018arx:deep} by using the cross-entropy method (CEM) \cite{deboer2005aor:cem} to select the best action. This process consists of randomly sampling $n_a$ actions from a multivariate Gaussian distribution (with diagonal covariance), evaluating their Q-values, selecting the best $n_b$ actions, refitting the sampling distribution to these actions, and iterating the procedure $n_n$ times until convergence, upon which the mean of the distribution is output as the best action.  Despite being an approximation, this approach works well in practice for low-dimensional action spaces.  (During training, $n_a = 16$, $n_b = 5$, and $n_n = 2$; during testing, $n_a = 64$, $n_b = 6$, and $n_n=3$.)

Training is performed entirely off-policy using a dataset of grasps collected with a pseudo-random policy that moves with a bias toward the object being grasped. This paradigm allows us to train and evaluate many different types of models without requiring new rollouts in simulation, which are time-consuming to collect.

\begin{figure}
\centering
\begin{tabular}{c}
\includegraphics[width=0.85\columnwidth]{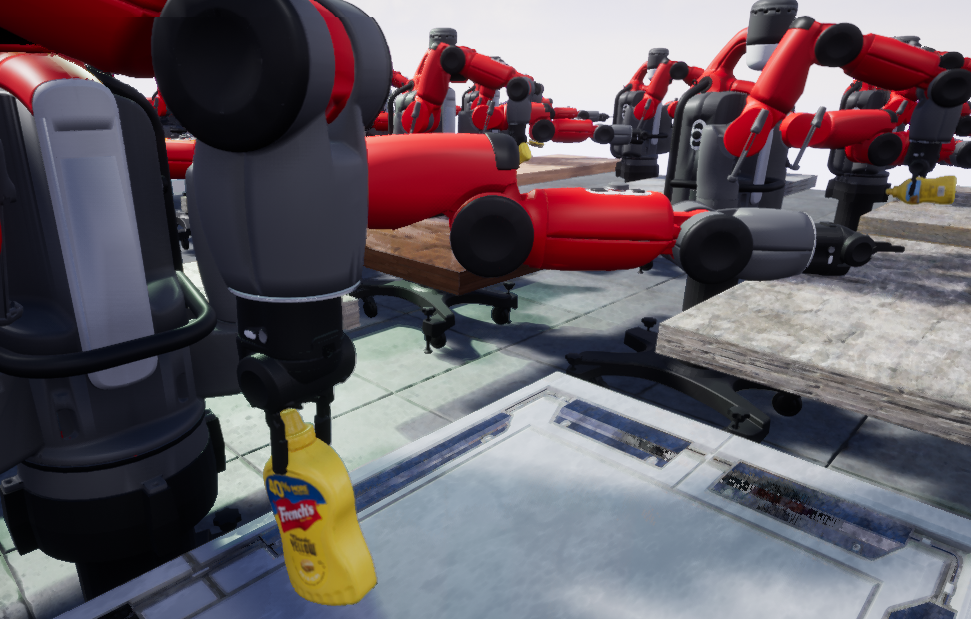} 
\end{tabular}
\caption{Domain randomization, including random table texture, is used to make the network robust to changes in the visual input.  Training time is reduced by running multiple simulated robots in parallel.}
\label{fig:baxter16_parallel_dr}
\end{figure}

During training, domain randomization~\cite{tobin2017iros:dr} is applied to bridge the reality gap.  
Specifically, random textures are applied to the table, as shown in Fig.~\ref{fig:baxter16_parallel_dr}, to assist the learned model to be invariant to various backgrounds and distractions.  
Similarly, each image's hue, saturation, and brightness is randomized. 
Additionally, we apply random Gaussian blur to the image (with kernel size uniformly selected from 1.0 to 3.0 pixels), in order to prevent overfitting to texture details in the object models that may vary from simulation to reality. 

During training, the initial position of the gripper is randomized to be within a $5 \times 5$~cm area at a height between $2$ and $4.5$~cm above the top of the object, with rotation offset from the desired object grasping position by as much as $\pm \pi / 4$~radians. 
These conditions were selected conservatively based on the error observed by our pose estimation algorithm~\cite{tremblay2018corl:dope}. 
From an initial position/orientation, the robot moves $k=5$ time steps before attempting to grasp and lift the object. 
The reward function gives a positive reward for a successful grasp at the end of an episode ({\em i.e.}, object is held by the gripper after the lift is performed), a negative penalty for distance to the centroid of the object, and a negative penalty for displacing the object prior to attempting a grasp. 

Note that, during training, the tips of the parallel-jaw gripper interact with the object according to the underlying physics engine.  
That is, the object is grasped and lifted only if there is sufficient force between the tips of the fingers to keep the object in place.
This is unlike previous work that either assumes that the object is grasped if the gripper is below some threshold in height \cite{quillen2018arx:deep}, or that uses analytic grasp calculations to estimate whether the grasp is successful \cite{mahler2017rss:dex}.
However, we discovered that the model learned to exploit a deficiency in the physics engine of our simulator, whereby it presses its gripper down on top of the object (rather than fit around it as desired), and the gripper would eventually snap into place around the object. 
As this behavior does not match the real world, due to the rigidity of the objects, we found it necessary to avoid this behavior by inserting an additional reward penalty on contact forces against the tips of the gripper's fingers prior to attempting a grasp.

\section{ROBOTIC SIMULATOR}

Two key components of a robotic simulator are its fidelity in modeling the physics of the world, and its fidelity in producing realistic sensory data.
Although existing physical simulators \cite{todorov2012iros:mujoco,rohmer2013iros:vrep,coumans2017:pybullet} have made significant strides toward both of these goals, there is much room for progress. 
In particular, photorealistic camera imagery and detailed contact modeling are two features that are missing from many off-the-shelf simulators.  
Below we describe our progress toward achieving these goals.
For an alternate approach, see~\cite{juliani2018arx:unity}, and for a comparison of various simulation tools, see~\cite{erez2015icra:simcompare}.

We have developed our own robotic simulator built on Unreal Engine 4 (UE4). 
UE4 is a highly customizable game engine with state-of-the-art rendering capabilities. 
UE4 has well established scene editing workflows and scripting tools, 
an active marketplace with a variety of 3D assets, 
procedural textures, rendering plugins, and a thriving developer community.
The simulator is agnostic to the underlying physics solver, with two options currently integrated: 
PhysX\footnote{\url{https://developer.nvidia.com/physx-sdk}} and FleX\footnote{\url{https://developer.nvidia.com/flex}}. 
PhysX models rigid bodies, whereas FleX is a GPU-based particle simulation library designed for real-time applications such as rigid body stacking, particle piles, soft bodies, and fluids. 
Both solvers allow the modeling of contact forces for object/robot interactions.

The simulator communicates with a controller using an efficient 
cross-application messaging system. 
This decoupling enables control code to switch between interacting with the simulator and real robot with minimal change.
The robot is controlled using either joint trajectories, velocities, 
accelerations, or forces through a configurable PID controller. 
Specific sensors such as force, depth images, RGB images, object 
segmentation, object poses, 2D bounding boxes, and 3D cuboids can be added/subscribed. 
At runtime, the scene can be modified such as changing the poses of 
object(s), camera(s) and/or robot(s).  
Physical objects in the simulator are represented using 
URDF which facilitates the introduction of new robots or objects. 
The simulator is designed for fast iterations and easy experimentation, by making the physics engines switchable at runtime, simplifying URDF import to dragging and dropping the file, and other facilities to emphasize ease of use.    
In addition, the simulator includes domain randomization tools such as randomizing 
textures, lights, colors, and object physical properties, to facilitate sim-to-real transfer.

\section{EXPERIMENTAL RESULTS}
\label{sec:results}

To evaluate the proposed approach, we performed a series of experiments both in simulation and in the real world.  The evaluation in simulation was performed on a held-out environment with lighting and textures remaining consistent throughout the experiments, as opposed to the highly randomized environment seen in the training data.  

All experiments were conducted with a Baxter robot (either in simulation or in reality).  
The Baxter has a simple parallel jaw gripper at the end effector, capable of traveling a total of approximately 4~cm between the two fingers. 
The built-in camera is mounted in the wrist aimed parallel to the fingers, producing $640 \times 400$~images at 30~Hz.

\subsection{Simulation}

\begin{figure*}
\centering
\begin{tabular}{cccc}
{\scriptsize \sf Meat Can} & {\scriptsize \sf Mustard Bottle} & {\scriptsize \sf Sugar Box} & {\scriptsize \sf Soup Can} \\
\includegraphics[width=0.475\columnwidth]{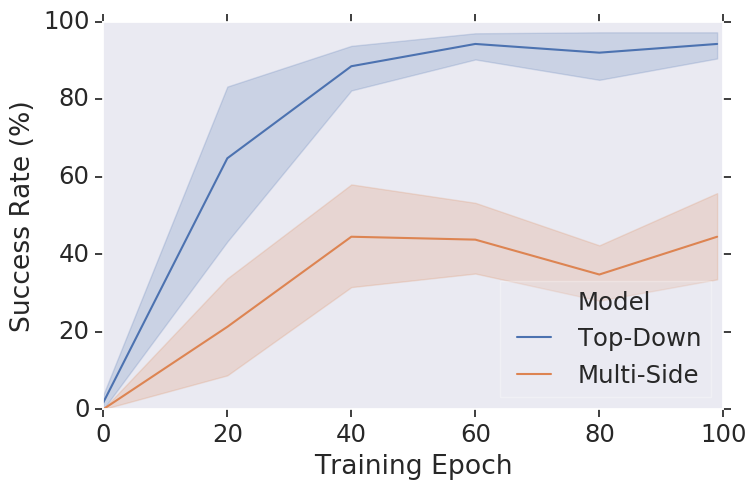} &
\includegraphics[width=0.475\columnwidth]{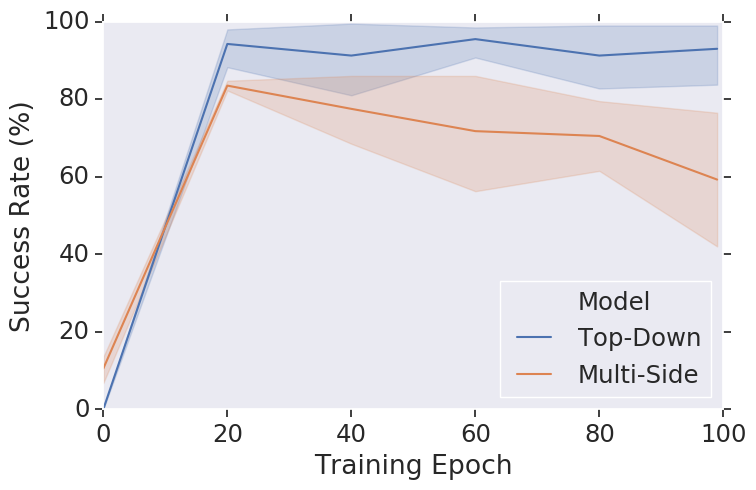} &
\includegraphics[width=0.475\columnwidth]{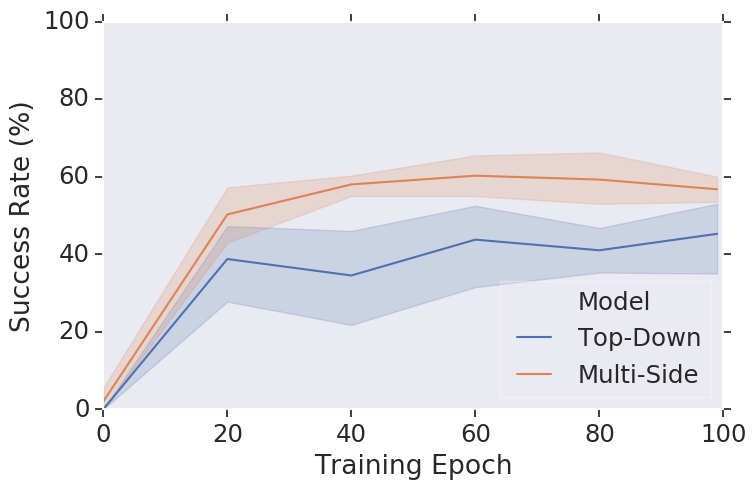} &
\includegraphics[width=0.475\columnwidth]{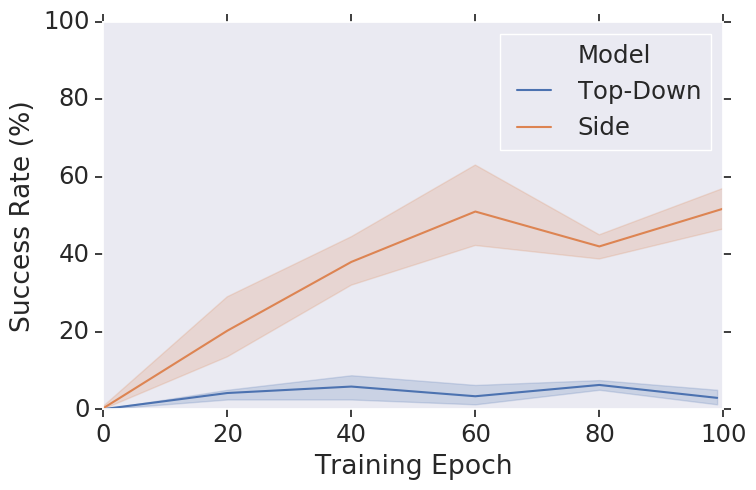}
\end{tabular}
\caption{Grasp success rate, evaluated on a simulated held-out environment, over the course of training for the four different YCB models.  Shown are results for top-down grasping as well as grasping from multiple sides (either the top or the two graspable sides).}
\label{fig:sim_results}
\end{figure*}

\begin{figure*}
\centering
\begin{tabular}{ccccc}
\includegraphics[width=0.360\columnwidth]{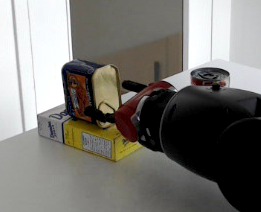} &
\includegraphics[width=0.360\columnwidth]{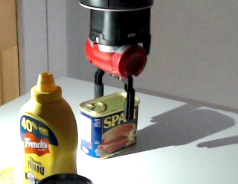} &
\includegraphics[width=0.360\columnwidth]{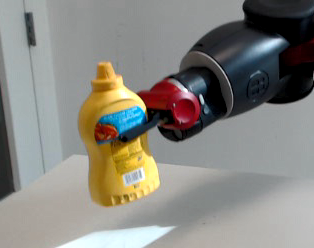} &
\includegraphics[width=0.360\columnwidth]{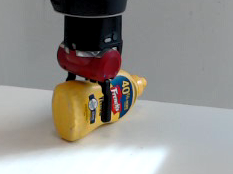} &
\includegraphics[width=0.360\columnwidth]{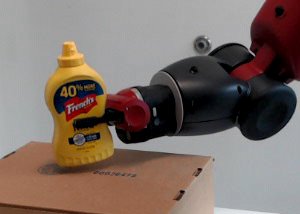} \\
\includegraphics[width=0.360\columnwidth]{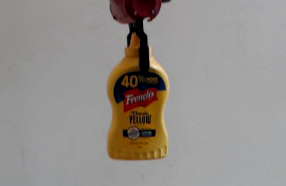} &
\includegraphics[width=0.360\columnwidth]{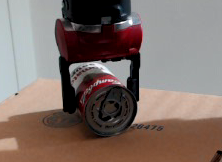} &
\includegraphics[width=0.360\columnwidth]{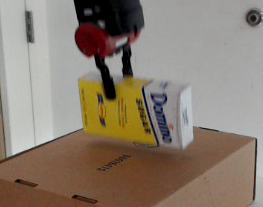} &
\includegraphics[width=0.360\columnwidth]{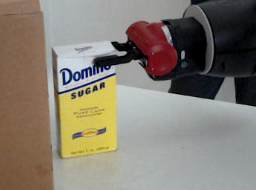} &
\includegraphics[width=0.360\columnwidth]{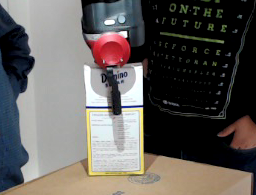} \\
\multicolumn{5}{c}{(a) Grasping of various objects from different directions by real Baxter robot.} \\ \\
\includegraphics[width=0.360\columnwidth]{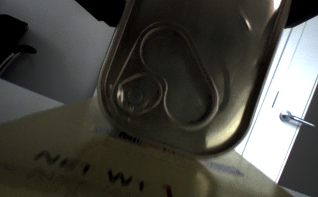} &
\includegraphics[width=0.360\columnwidth]{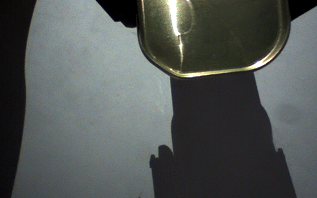} &
\includegraphics[width=0.360\columnwidth]{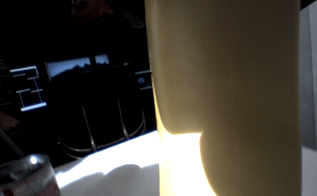} &
\includegraphics[width=0.360\columnwidth]{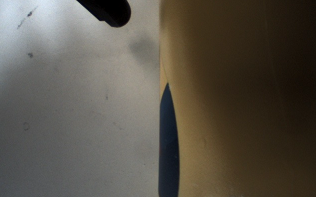} &
\includegraphics[width=0.360\columnwidth]{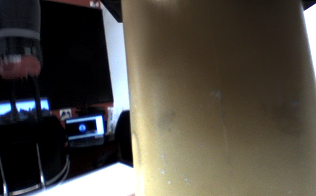} \\
\includegraphics[width=0.360\columnwidth]{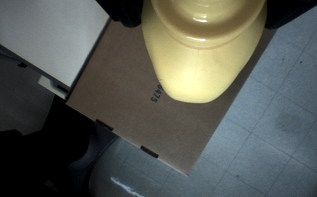} &
\includegraphics[width=0.360\columnwidth]{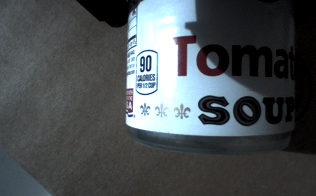} &
\includegraphics[width=0.360\columnwidth]{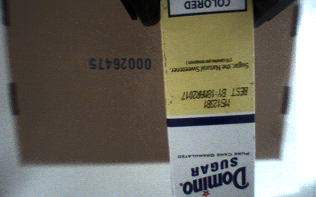} &
\includegraphics[width=0.360\columnwidth]{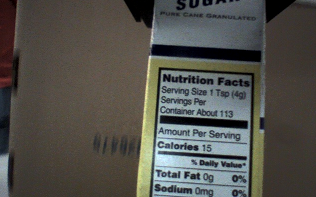} &
\includegraphics[width=0.360\columnwidth]{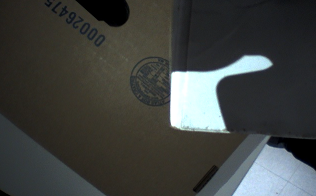} \\
\multicolumn{5}{c}{(b) Corresponding images from the wrist-mounted camera.} 
\end{tabular}
\caption{After training in simulation, the policy is able to grasp the intended object in the real world without fine-tuning.  Because the action space is gripper-centric, the robot is capable of grasping from various orientations, such as vertical and horizontal.  Note the variety of top-down and side grasps, as well as lighting conditions and backgrounds.}
\label{fig:real_exp1}
\end{figure*}

We used the procedure described earlier to train a network using four different objects from the YCB dataset \cite{calli2015icar:ycb,calli2015ram:ycb}: meat can, soup can, sugar box, and mustard bottle. 
These objects were chosen to cover a variety of geometric shapes (rounded cuboid, cylinder, squared cuboid, and a unique curved shape), material properties (matte, reflective), and colors (brightly colored, dark).  
They were also selected because detection/pose estimation networks for them are already available \cite{tremblay2018corl:dope}, which can automatically provide initial pre-grasp locations (although not used in this work).
For all objects, two models were trained:  one for grasping the top of the object (top-down) and another for grasping from all directions (multi-side).
More specifically, for all but the soup can, the multi-side models were trained with data from all three directions (top and the two graspable sides) so that they are capable of grasping from either the top or two side directions; whereas the soup can was trained to grasp from any of its infinite side grasping directions.  
All models were trained with 100k data samples each.

Training results are shown in Fig.~\ref{fig:sim_results}.  
Both the meat can and mustard bottle achieved 95\% success or more with the top-down grasp models, and moderate success at multi-side grasping.  
With the sugar box, results are inverted, with the multi-side version performing better than top-down.  
This inversion is due to the lack of texture on the white top of the box, coupled with the fact that the held-out environment used for evaluation consisted of a light-colored table.
The soup can achieved promising success for side grasping, but top-down grasping proved to be challenging in simulation due to the gripper's tendency to slip off the curved surface from incorrect friction modeling.

An ablation study was performed to evaluate the effectiveness of specific components on successful training.
These components were isolated and removed, and the resulting performance was then measured. 
We found that, without image augmentation, the model improves quickly but then reaches a maximum, after which it declines in performance due to overfitting.  
Behavior without dropout was similar, with an even more pronounced effect. 
Similarly, removing auxiliary targets decreased overall performance slightly while increasing training time. 
Thus image augmentation, dropout, and auxiliary targets are important for good results.

\subsection{Real world}

To quantify the performance in the real world, we placed the YCB objects at various places on a table.  
Due to the poor quality of the in-hand Baxter camera, we had to manually adjust the gain of the camera; and the room lighting was also adjusted to reduce the effects of reflections that are not modeled in simulation.
The grasp success rate, starting from a manual pre-grasp above the object in roughly the same area as in simulation, is shown in Tab.~\ref{tab:realworld}.
These results show that the network successfully learned policies that transfer to the real world under challenging conditions.
As expected, the soup can performs better on side grasps than top grasps.  
The mustard bottle is the reverse, with top grasps being easier than side grasps.
Note that top-down grasping of the soup is better in the real-world than simulation due to the increased friction in the real world that yields more forgiving behavior when the can is grasped off-center.

\begin{table}
\centering
\caption{Grasp success rate of various objects in the real world, after training only in simulation.}
\begin{tabular}{c|c|c}
 & \multicolumn{2}{c}{grasp success (\%)} \\
object & top & multi-side \\
\hline
meat can & 83.3 & 75.0 \\
mustard bottle & 100.0 & 66.7 \\
sugar box & 50.0 & 66.7 \\
soup can & 33.3 & 50.0 
\end{tabular}
\label{tab:realworld}
\end{table}

Fig.~\ref{fig:real_exp1} shows examples of the policy learned in simulation successfully grasping objects in the real world.  
The system is capable of grabbing the four objects on which it was trained from various directions (side and top).
Note the variety of lighting conditions, including dark and bright images, with and without specular highlights.
Because the YCB object models lack material properties, the metallic top of the meat can causes specular reflections in reality that are not present in the simulated images, as also noted in \cite{tremblay2018corl:dope}.  
Though our approach is able to ignore small reflections, more accurate models in simulation are needed to alleviate this problem.

\section{PREVIOUS WORK}
\label{sec:previous}

Relationship to previous work is considered in this section.

\textbf{Robotic grasping} is fundamental to pick-and-place tasks and has been studied by a number of researchers.
For example, Dex-Net~\cite{mahler2016icra:dex,mahler2017rss:dex,mahler2018icra:suction,mahler2017corl:binpicking} is a deep network aimed at \emph{indiscriminate} top-down grasping of objects in a bin using an overhead depth camera; the network is trained entirely in simulation.
In contrast, the approach of Levine~\etal\cite{levine2018ijrr:learning} learns an indiscriminate top-down grasp prediction model from an overhead RGB camera using hundreds of thousands of grasp attempts by a collection of real robots.
Jang~\etal~\cite{jang2017corl:endtoend} address the problem of \emph{semantic grasping}, in which a two-stream network learns top-down grasps of specific classes of objects, trained on tens of thousands of manually-labeled real RGB images and hundreds of thousands of images utilizing self-supervised label propagation.
Morrison~\etal\cite{morrison2018rss:grasp} learn a mapping from a wrist-mounted depth image to the quality and pose of grasps at every pixel, trained on labeled real depth images.
Morrison~\etal\cite{morrison2018icra:cartman} combine RGB pixelwise semantic segmentation with depth sensing for multimodal top-down semantic grasping.
Quillen~\etal\cite{quillen2018arx:deep} compare a variety of off-policy deep RL methods for indiscriminate top-down grasping using purely simulated data for both training and test.
Our method can be seen as an extension of these works by introducing the problem of \emph{directional semantic grasping}.

\textbf{Domain adaptation} aims to allow a network trained on simulated data to generalize to real data, using some amount of labeled real data.  Various researchers have explored ways to reduce the amount of labeled real data needed.
Fang \etal \cite{fang2017arx:multi} address top-down grasping of household dishes from instance-segmented images from a wrist-mounted RGB camera; sim-to-real transfer is accomplished using multi-task domain adaptation with real indiscriminate grasping data.
Bousmalis~\etal\cite{bousmalis2018icra:drg} present a GAN-based domain adaptation model for sim-to-real transfer for top-down indiscriminate grasping using a monocular RGB camera.
Zhang \etal\cite{zhang2017arx:adversarial} propose an adversarial discriminative transfer approach to reduce the amount of labeled real data required for sim-to-real transfer for the task of reaching a blue cube using an RGB camera and an end-to-end network.
Sadeghi \etal\cite{sadeghi2018cvpr:sim2real} use an auxiliary adaptation loss to fine-tune the perception layers of a neural network using a small amount of annotated real data, to solve the problem of viewpoint-invariant visual servoing to reach a desired object.
In contrast with this work, we build upon the success of our earlier work~\cite{tremblay2018corl:dope} by relying solely upon domain randomization, without domain adaptation; this requires the simulator to produce data that approximates the real distribution.

\textbf{Sim-to-real transfer without domain adaptation} is the ambitious goal of training in simulation and executing in the real world without any labeled real-world data.
James \etal\cite{james2017corl:transferring} learn an end-to-end policy entirely in simulation to grasp a red cube and place it inside a blue basket using RGB images; sim-to-real transfer is accomplished via domain randomization.  
Similarly, Matas \etal\cite{matas2018arx:sim} train an end-to-end deep RL policy in simulation to manipulate deformable objects using RGB images; domain randomization is used to cross the reality gap.
Yan~\etal\cite{yan2017nipsw:simtoreal} use imitation learning to grasp a tiny yellow sphere from RGB images by training a network to perform a binary segmentation of the sphere from the background.
In all this previous work, the object to be grasped is a single, solid color; our work can be seen as an extension of such work to graspable objects with more realistic, textured appearance. 

The transfer is easier when performed purely with depth images, although this prevents the network from using valuable visual information, and it requires less standard hardware with additional error modes.  
Vierech \etal\cite{vierech2017corl:visuomotor} learn a visuomotor controller for indiscriminate top-down grasping using a wrist-mounted depth sensor, trained in simulation.
Osa \etal\cite{osa2016iser:hrl} use hierarchical RL to autonomously collect a simulated grasping dataset for learning grasping strategies for isolated objects from depth images; the system is capable of grasping from multiple angles.
Fang~\etal\cite{fang2018rss:tog} use self-supervision to generate simulated data for learning task-oriented grasping for tool manipulation from depth images.
In contrast, our approach works solely with RGB images.

\textbf{Object manipulation} broadly encompasses pick-and-place tasks of known objects in the robot workspace.
The MOPED framework~\cite{collet2011ijrr:moped} performs real-time detection and pose estimation of multiple known objects from one or more RGB images for the purpose of object manipulation.
SimTrack~\cite{pauwels2015iros:simtrack} is a framework for real-time detection and tracking of multiple known objects using combined RGBD/RGB (overhead/in-hand) sensors, for the purpose of grasping, manipulating, and stacking objects in a tabletop environment.
Sui \etal\cite{sui2017ijrr:goaldir} address the problem of axiomatic scene estimation for goal-directed top-down manipulation of rigid objects on a table, using an RGBD sensor.
Zeng \etal\cite{zeng2018icra:semantic} use a 2D object detector and depth sensing to find the poses of known objects in a tabletop scene using an RGBD sensor, and motion planning for grasping and manipulation.
Our work is an investigation into using reinforcement learning to solve a subproblem, namely grasping, within this larger context.

Since our goal is also sim-to-real transfer of an end-to-end policy trained on simulated RGB images without any labeled real data, our work is most similar to that of James \etal\cite{james2017corl:transferring} and Matas \etal\cite{matas2018arx:sim} above.  
Like their approaches, we also use domain randomization to bridge the reality gap.  
Our work differs in our use of a simulator that models physical contact and generates photorealistic imagery for learning to map images of textured objects to actions that aim to grasp specific objects from specific angles.

\section{DISCUSSION AND CONCLUSION}

In this paper we have attempted an ambitious goal, namely, sim-to-real transfer of a closed-loop control policy using a single RGB image as input, and using an off-the-shelf deep reinforcement learning (RL) algorithm (namely, DDQN) without any additional supervision.
The fact that the learned policy transfers to the real world at all is promising.
We have shown successful grasps of all objects on which the system was trained, from various directions and in challenging conditions.
For training, we developed a robotic simulator that models physical reality (including contact) and is capable of generating photorealistic and domain randomized training data. 

Although our set of objects may seem small, it should be kept in mind that there simply do not exist (to our knowledge) large-scale datasets of household objects with textured 3D models.
Existing datasets of 3D models like ShapeNet~\cite{chang2015arx:shapenet} or Pascal3D+~\cite{xiang2014wacv:pascal3d} are not textured.
With such non-textured models, it is possible to train algorithms in simulation that transfer to reality using depth cameras~\cite{mahler2017rss:dex,danielczuk2016icra:mechsearch}, since depth is not affected by texture; but transfer using RGB images is not likely to succeed.
An alternative is to use real images for training~\cite{jang2017corl:endtoend}, but these are expensive and time-consuming to gather, even for top-down grasping; it is not straightforward to extend such an approach to directional grasping, as in our work.

Although the YCB dataset~\cite{calli2015icar:ycb} does include more objects than the subset we use here, many of these objects are not suitable for grasping experiments, due to safety (scissors), lack of texture (wooden block), size (the coffee can is too big for the Baxter gripper), or availability (the texture of the 3D gelatin model in YCB no longer matches the current product sold at stores), and so forth---thus making sim-to-real experiments impossible.  
The four objects selected in this work are representative of other objects in this category of household objects, thus suggesting that the results presented here would transfer to similar objects.

Despite the impressive success of deep RL in other areas, such as mastering games like Atari~\cite{mnih2015nature:dqn} and Go~\cite{silver2016mastering}, fundamental challenges remain to apply such algorithms to robotics.  
These challenges include (among other things) sample inefficiency, slow convergence, and sensitivity to initial conditions and hyperparameters.
Our results echo the findings of other researchers~\cite{james2016nipsw:simrobarm}, namely, that sim-to-real transfer using deep RL is extremely difficult and not yet possible in a robust manner.
For this reason, much success to date has still relied upon supervised learning rather than RL~\cite{james2017corl:transferring}.
Given such success, we believe a supervised learning approach would be worth exploring for the problem of directional semantic grasping addressed here.

We hope the promising results of our approach will encourage additional research in this area.
Over time, robotic simulators such as ours will continue to improve in their modeling ability, thus further reducing the reality gap.
Future work should explore ways to improve the robustness of the proposed method by 
expanding the set of objects (to include, for example, thin and irregularly shaped objects), generalizing to unknown objects, utilizing supervised learning, addressing extreme lighting conditions, improving contact and friction modeling, and including temporal information in the processing.

\section*{ACKNOWLEDGMENTS}

We thank Stephen Tyree, Iuri Frosio, Ryan Oldja, and Abhishek Raj Dutta for their help with the project.

\bibliographystyle{IEEEtran}
\bibliography{icra2020-shariq}

\end{document}